# Path Planning for Multi-Copter UAV Formation Employing a Generalized Particle Swarm Optimization

**Van Truong Hoang**
**Naval Academy, Vietnam**
Email: vantruong.hoang@alumni.uts.edu.au

**Abstract**
The paper investigates the problem of path planning techniques for multi-copter uncrewed aerial vehicles (UAV) cooperation in a formation shape to examine surrounding surfaces. We first describe the problem as a joint objective cost for planning a path of the formation centroid working in a complicated space. The path planning algorithm, named the generalized particle swarm optimization algorithm, is then presented to construct an optimal, flyable path while avoiding obstacles and ensuring the flying mission requirements. A path-development scheme is then incorporated to generate a relevant path for each drone to maintain its position in the formation configuration. Simulation, comparison, and experiments have been conducted to verify the proposed approach. Results show the feasibility of the proposed path-planning algorithm with GEPSO.

**Abbreviations**
    PSO         Particle Swarm Optimization
    GEPSO       Generalized PSO
    UAVs        Unmanned Aerial Vehicles

## 1. Introduction

Drones, a class of UAVs, have contributed to many areas in the practical world. Their investigations have been broadening while exploiting computing and sensing technology development. Compared to only one UAV, a group of drones can perform more problematic tasks at increased powers, so UAV cooperation in formation is growing great interest both in research and the practical world. Among those studies, path planning recreates a vital role for drones to move freely in tough environments with several constraints in formative manners [1]. The path generating problem for drone formation is also complex to achieve an optimal path and meet the constraints of specific missions.

There have been many examinations on this topic, with many algorithms presented. These approaches were A*, D*, RRT (Rapidly Exploring Random Tree) or, PRM (Probabilistic Roadmap Method) [2-4], and machine learning [5]. Their advantages include creating an appropriate flyable trajectory and obstacle collision avoidance capability. However, the resulting paths may not meet optimization requirements or demand extensive training data.

The literature recently paid attention to nature-inspired algorithms such as GA, ACO, TLBO, and PSO (Genetic Algorithm, Ant Colony, Teaching-learning-based, and Particle Swarm Optimizations, respectively), which have been devoted to the mentioned problems. These techniques satisfy the UAVs' flyability, feasibility, and optimal paths. GA can solve mixed path-planning problems [6]. ACO is strong in minimization of path length and collision avoidance [7]. TLBO is a simple implementation with a lower convergence rate [8]. However, their weakness is that they are not always guaranteed the condition of optimal paths and have a non-decisive nature in presenting solutions. In 3D-space path planning, they face different conflicts, i.e., GA requires heavy computation and has premature convergence, ACO has a low convergence rate, and TLBO is unsuitable for complicated working spaces [9].

Thanks to the flexibility of its parameters and concepts, the PSO-based algorithm has been investigated in a broad range of applications. Several PSO variants have been successfully applied in the UAV path planning problem in 3D spaces [10-12]. However, producing a path in a complicated space with many obstacles is still problematic, particularly when PSO-based algorithms are vulnerable to optimization execution in case of the mistaken choice of coefficients [13]. The Generalized Particle Swarm Optimization (GEPSO) [14] enhances the original by increasing connections among particles by sharing their knowledge. The GEPSO increases the swarm's diversity and provides powerful examination capabilities in unknown environments. Besides, the inertial weights are dynamically updated and controlled to accelerate the convergence rate. The above properties make GEPSO capable of solving problems in complex operational spaces. Therefore, in this work, GEPSO will be utilized to solve the path planning problem for multi-UAV formation with the capability of finding reference paths that satisfy the safe, dynamic, and task undertaking requirements in a multi-constraint cost function.

Our work starts with the problem of a multiple drone formation working in an intricate 3D environment. The examined working space and its attributes are obtained from a satellite map. Therein, an extra multi-objective fitness function is created to simultaneously (1) generate the shortest path, (2) improve collision avoidance power, and (3) guarantee safety and task efficiency. We then investigate the capability of GEPSO in the global optimal solution to generate a designed path for the formation centroid while fulfilling the above objectives. The resulting path is finally translated into the desired route for each UAV to preserve the formation shape. Different from path planning for UAV formation [15], the advantages of the proposed approach are feasible and capable of solving the problem in an intricate working space, generating an optimal, flyable and feasible path for each UAV in the formation while preserving the surface inspection mission. To illustrate the validity and effectiveness, we have conducted several experiments and comparisons. The results depict the outperformance of the proposed methods.

The paper is arranged as follows. Section 2 states the multi-objective path planning problem for the formation. Section 3 describes the GEPSO algorithm. Section 4 demonstrates the implementation of the path planning algorithm for the formation and then each UAV. Simulation and experimental results are expressed in Section 5. The conclusion and future work will be the end of this paper.

## 2. Problem statement

### 2.1 Path planning for a UAV's formation

A single UAV's trajectory is formed from the moving components of the UAV's center of gravity. The 3D motion of UAVs consists of six main components: three translationals and three rotationals. A flight path is usually determined by a transition point (waypoint) intersecting two consecutive lines, so we have a set of segments connected from starting to ending points. Waypoints and movements are defined in the Cartesian coordinate system ($Oxyz$). The UAV state is defined as $P(x, y, z, \theta, \psi)$, where $(x, y, z)$ is the position, $\theta$ and $\psi$ are the horizontal and vertical rotation angles, respectively. However, multi-copters are usually flexible enough to perform rotation angles without limitation, so only the three translational movements represent the state of the UAV.

When a UAV moves from the starting state $P_s$ to the ending state $P_f$ then the problem of establishing a flight trajectory includes creating one or more flight generated paths $r(q)$ connecting $P_s$ and $P_f$. Mathematically, this can be represented as:

$$P_s \xrightarrow{r(q)} P_f \quad (1)$$

where $q$ is defined as the path parameter.

For a UAV fly from a location with start state $P_s(x_s, y_s, z_s)$ to a location with final state $P_f(x_f, y_f, z_f)$, (1.1) can be expressed as:

$$P_s(x_s, y_s, z_s) \xrightarrow{r(q)} P_f(x_f, y_f, z_f) \quad (2)$$

Extending (2) to $N$ UAVs in a rigid body formation, we have:

$$P_s(x_s^i, y_s^i, z_s^i) \xrightarrow{r(q)} P_f(x_f^i, y_f^i, z_f^i), i = 0, \cdots, N. \quad (3)$$

However, the path created may be unusable because the UAV cannot perform instantaneously change its state at each waypoint. In practice, many constraints are related to path planning problems, most specific to UAVs. The two most important are feasible and safe. A feasible trajectory must meet motion constraints. The safety is achieved by avoiding obstacles in space limited. In addition, there may be other constraints to fulfil purposes of flying.

We use the symbol $\coprod$ to represent constraints, so the problem of path planning with constraints can be expressed as [10]:

$$P_s(x_s^i, y_s^i, z_s^i) \xrightarrow{\coprod r^i(q)} P_f(x_f^i, y_f^i, z_f^i) \quad (4)$$

### 2.2 Formation model

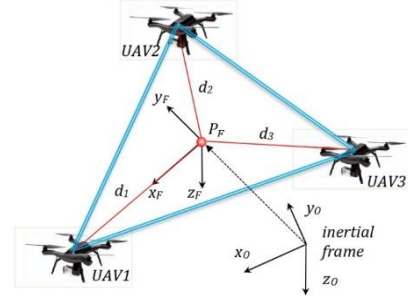

**H. 1**  *Working space acquisition*

In our work, the selected shape of the formation is a rigid triangle with three UAVs, as illustrated in H. 1. The formation travels in two main coordinate systems, i.e., the body frame $\{x_F, y_F, z_F\}$ and the fixed inertial frame $\{x_O, y_O, z_O\}$. The moving formation body frame is defined such that the origin $P_F$ coincides with the triangle's centroid, the $x_F$ axis is directed from the centroid of the triangle to the UAV1 (head), the $z_F$ axis is perpendicular to the plane containing UAVs and directed downward, and the $y_F$ axis is perpendicular to the $(x_F, z_F)$ plane. Placements of the $n$th UAV, $n = 1, 2, 3$, in $\{x_O, y_O, z_O\}$ are implied as $P_n = \{x_n, y_n, z_n\}$. The body frame allows for determining orientation concerning the fixed inertial frame. The centroid of the shape represents the formation location as:

$$P_F = \frac{1}{3} \sum_{n=1}^{3} P_n \quad (5)$$

where $R_{OF}$ is the rotation matrix that represents the relation between $\{x_F, y_F, z_F\}$ and $\{x_O, y_O, z_O\}$.

To generate the specific trajectory for the $n$th UAV, we represent $T_n = T_F + \Delta T_n$ with $T_F$ as the path generated by the planning algorithm and $\Delta T_n$ as the required trajectory change for the $n$th UAV to shift it away from the group's centroid. This difference in position is calculated based on the desired relative distances among the UAVs and the relative position errors, $\Delta T_n = [e_{n,x}, e_{n,y}, e_{n,z}]^T$ with $[e_{n,x}, e_{n,y}, e_{n,z}]$ is the relative vector of distances from the centroid in the inertial frame. The output $T_n$ will be the trajectory command used by the UAV onboard controllers for trajectory tracking.

### 2.3 Multi-objective Cost Function
*a. General cost function*

Setting up a reasonable cost function is extremely important when applying optimization algorithms to solve practical problems. The cost function usually includes at least a length constraint and a constraint on avoiding obstacles that threaten the safety of the established flight path. In monitoring missions, other constraints that the UAVs must meet include limitations on flight altitude and maintenance of distance to monitoring objects. Altogether, we can combine the individual costs into a common function, as follows:

$$\amalg r(q) = \amalg_{range} r(q) + \amalg_{safe} r(q) + \amalg_{alt} r(q) + \amalg_{mission} r(q) \quad (6)$$

where $\amalg_{range} r(q), \amalg_{safe} r(q), \amalg_{alt} r(q)$, and $\amalg_{mission} r(q)$ are the specified criteria for path length, obstacle avoidance, flight altitude and monitoring range.

*b. Flight path length cost*

For a given flight path, with a total number of waypoints *n*, the coordinates of the waypoints can be expressed as:

$$x_1^i, x_{n+1}^i, x_{2n+1}^i \, , \, x_2^i, x_{n+2}^i, x_{2n+2}^i \, , \cdots, \, x_n^i, x_{2n}^i, x_{3n}^i$$

or

$$W_1^i, W_2^i, \cdots, W_n^i$$

The path length cost $\amalg_{range} r(q)$ is defined as the sum of all path lengths from the start point to the endpoint:

$$\amalg_{range} r(q) = p_{range} \sum_{j=0}^{n+1} \left\| \overrightarrow{W_j, W_{j+1}} \right\| \quad (7)$$

where $W_0$ and $W_{n+1}$ are the starting and ending points of the trajectory segment, respectively, the symbol $\|.\|$ represents the Euclidean distance of a vector. $W_0$ and $W_{n+1}$ do not change because all particles have the same starting and ending points.

*c. Obstacle avoidance cost*

We have a set of *K* areas where dangerous signs exist for UAVs in the operating area, collectively called obstacles and are denoted $\{T_1, T_2, \ldots, T_K\}$. Circles with center $C_k$ and different radii $R_k$ represent these regions. The obstacle's position is located at the center, and its radius indicates the coverage of this area.

The distance from a given flight path segment $\overrightarrow{W_j, W_{j+1}}$ to the center of each obstacle is calculated according to its midpoint to the obstacle center. To ensure a safe distance, a UAV is only allowed to fly near a point outside the range of the obstacle. The value of the obstacle avoidance function will be calculated if the UAV flies into the cylinder ($C_k$, $R_k$). The detailed implementation of the calculation method of the hazard index function value can be described as follows:

Step 1. For each obstacle $T_k$, calculate the distance from its center $C_k$ to the projection of the flight path segment $\overrightarrow{W_j, W_{j+1}}$, denoted as $d_k$.

Step 2. Compare the size $d_k$ and radius of the cylinder $R_k$. If $d_k \geq R_k$, the $\amalg_{safe,k}$ function value of the *k*th obstacle for the segment ($\overrightarrow{W_j, W_{j+1}}$), is 0, that is:

$$\amalg_{safe,k} \overrightarrow{W_j, W_{j+1}} = 0 \quad (8)$$

Otherwise ($d_k < R_k$), move to Step 3.

Step 3. Calculate the length of projection $\overrightarrow{W_j', W_{j+1}'}$ covered by the *k*th obstacle, denoted as $l_k$, we then have safe cost of the segment to *k*th obstacle:

$$\amalg_{safe,k} \overrightarrow{W_j', W_{j+1}'} = \begin{cases} R_k l_k & \text{if } d_k \leq l_k \\ R_k l_k / d_k & \text{if } d_k > l_k \end{cases} \quad (9)$$

Step 4. The obstacle avoidance cost function of the entire flight path can be expressed as follows:

$$\amalg_{safe} r(q) = p_{safe} \sum_{k=1}^{K} \sum_{j=0}^{n+1} s_k \amalg_{safe,k} \overrightarrow{W_j, W_{j+1}} \quad (10)$$

where *K* is the total number of obstacles and $s_k$ is the danger level of the *k*th obstacle.

*d. Altitude limitation cost*

For UAVs, flying at a limited altitude is a critical factor to improve the mission's effectiveness. When flying over terrain, the UAVs must fly over terrain at a minimum altitude $z_{min}$ to prevent ground crashing. The UAVs must fly below the maximum altitude $z_{max}$ to increase the visibility probability of monitoring. We use the symbol of terrain elevation at the reference point $W_{ij}$ as $T_{ij}$. Then, the relative altitude $h_{ij}$ of the UAVs at this point is the difference between the absolute altitude $x_{i,2n+j}$ and the terrain altitude $T_{ij}$:

$$h_{ij} = x_{2n+j}^i - T_{ij}$$

As illustrated in H. 6, the flight altitude cost $\amalg_{alt} r(q)$ can be determined as follows:

$$\amalg_{alt} r(q) = p_{alt} \sum_{j=0}^{n} dh_{ij}$$

$$dh_{ij} = \begin{cases} h_{ij} - z_{max}, & \text{when } h_{ij} > z_{max} \\ 0, & \text{when } z_{min} \leq h_{ij} \leq z_{max} \\ z_{min} - h_{ij}, & \text{when } 0 \leq h_{ij} < z_{min} \\ \infty, & \text{when } h_{ij} \leq 0 \end{cases} \quad (11)$$

where $p_{alt} > 0$ is the penalty coefficient.

*e. Mission performance cost*

To enhance the usefulness of onboard monitoring sensors, UAVs must fly within a certain distance of the monitoring object(s):

$$d_n^s \in \left[ d_{min}^s, d_{max}^s \right], \quad (12)$$

where $d_n^s, d_{min}^s$, and $d_{max}^s$ represent the current, minimum, and maximum distances from $UAV_n$ to the object, respectively. The mission execution cost of the UAV is then expressed as:

$$\amalg_{mission,q} r(q) = \begin{cases} 0 & \text{if } d_{min}^s \leq d_n^s \leq d_{max}^s \\ d_{min}^s - d_n^s & \text{if } d_n^s < d_{min}^s \\ d_n^s - d_{max}^s & \text{if } d_n^s > d_{max}^s \end{cases} \quad (13)$$

The mission cost function of the entire formation can be expressed as follows:

$$\amalg_{mission} r(q) = p_{mission} \sum_{n=1}^{N} \sum_{q=1}^{Q} \amalg_{mission,q} r(q) \quad (14)$$

where $Q$ is the total number of waypoints, $p_{mission}$ is the mission performance penalty.

## 3. Proposed path planning algorithm
### 3.1 Particle swarm optimization

Particle swarm optimization is an evolutionary computing technique developed in 1995 [16]. PSO is inspired by and based on research on the social behavior. It uses a number of individuals (particles) flying through the hyperspace (working space) of the problem at a given speed. At each iteration, the velocities of individual individuals are randomly adjusted according to the historical best position for that individual and the neighboring best position. Each individual's movement naturally evolves to an optimal or near-optimal solution. PSO has some outstanding advantages compared to other optimization methods, i.e., (i) more accessible with fewer parameters to adjust, (ii) highly productive, and (iii) more effective in maintaining flock diversity [13].

In PSO, each feasible solution can be modeled as an individual moving in the hyperspace of the problem. The position of each individual is determined by the vector $x_i \in R^n$ and its motion is characterized by the velocity of the individual $v_i \in R^n$, as follows:

$$x_i^t = x_i^{t-1} + v_i^t, \quad (15)$$

The information available to each individual is their own experience and knowledge of the others' performance. Since the relative importance of these two factors may vary from decision to decision, it is reasonable to apply a random weight to each part. Thus, the velocity will be determined by:

$$v_i^t = \omega v_i^{t-1} + \varphi_1 r_1 \left[ p_i^{t-1} - x_i^{t-1} \right] + \varphi_2 r_2 \left[ p_g^{t-1} - x_i^{t-1} \right], \quad (16)$$

in which, $\varphi_1$ and $\varphi_2$ are two positive numbers, $r_1$ and $r_2$ are two random numbers with uniform distribution in the range [0 1].

### 3.2 Generalized particle swarm optimization

Because of its outstanding features, PSO has been a promising and effective optimization method for multi-objective optimization problems. Significant modifications have been made to enhance the original performance, such as discrete PSO, θ-PSO and hybrid PSO. However, these improvements are based on the position and velocity framework and do not deviate from the usual position and velocity update rules. These proposals typically involve changes to the PSO update equations without changing its structure.

The Generalized PSO [14] was proposed to overcome the abovementioned issues. GEPSO is powerful in the interrelations among particles, accelerating the swarm convergence. Some random velocities creates a better exploration in various complicated search spaces. In GEPSO, the position equation remains the same as in its original, while the velocity is adjusted by some new terms, as below:

$$v_i^t = \psi \left( a_1 v_i^{t-1} + a_2 \left[ p_i^{t-1} - x_i^{t-1} \right] + a_3 \left[ p_g^{t-1} - x_i^{t-1} \right] \right. \\ \left. + a_4 \left[ p_{rand}^{t-1} - x_i^{t-1} \right] + a_5 v_{rand}^{t-1} \right) \quad (17)$$

where $\psi$ is the constriction parameter, $a_1 = \omega_1^i$, $a_2 = \omega_2 \varphi_1 r_1^i$, $a_3 = \omega_3 \alpha_1 \varphi_2 r_2^i$, $a_4 = \omega_4 \alpha_2 \varphi_3 r_3^i$, and $a_5 = \omega_5 \alpha_3 \varphi_4 r_4^i$, $\alpha_1, \cdots, \alpha_4$ are regulated probability constants, $\omega_1, \cdots, \omega_5$ are inertia weight factors, and $\varphi_1, \cdots, \varphi_4$ are some acceleration constants.

The updated formula for $\psi$ is represented as:

$$\psi = \frac{2}{\left| 2 - \varphi_2 - \varphi_3^2 - 5\sqrt{\varphi_2 + \varphi_3} \right|}, \quad (18)$$

The below equation is used to update $\omega_1$ at each function evaluation dynamically:

$$\omega_1 = \min\{\omega_{\min}, \omega_\kappa\} \\ \omega_\kappa = \omega_1^{t-1} - \left[ \frac{\omega_{\max} - \omega_{\min}}{N} i \left[ f(p_g^{t-1}) - f(p_g^{t-2}) \right] \right], \quad (19)$$

where $\omega_{\max}, \omega_{\min}$ are the maximum and minimal inertia value, respectively; $i, N$ in order are the current and total number of iterations; $f(p_g^{t-1}) - f(p_g^{t-2})$ is the previous function evaluation value.

The initial values of positions and velocities of the particles are calculated as:

$$x_i^0(x_k) = x_{k,\min} + \text{rand} \cdot x_{k,\max}, \quad (20)$$

$$v_i^0(x_k) = x_i^0(x_k) + \gamma_k, \quad (21)$$

where $x_{k,\min}$ and $x_{k,\max}$ are the lowest and highest values of the $k$th coordinate of the particles, respectively.

## 4. Path planning implementation
### 4.1 Preparation

The preparation begins when the UAV group is selected for a monitoring task. Required tasks include (1) deciding the working zone, (2) analyzing the working space to find terrain and obstacles data, and (3) deciding the flying range of altitude as well as the set of mission performance distance to the object. By analyzing the mission, the take-off ($P_s$) and landing ($P_f$) waypoints are first determined. In this work, the working space is determined by using a rectangular box defined by the four GPS latitude ($\varphi_i$) and longitude ($\lambda_i$) coordinates, $\Gamma_i = \{\varphi_i, \lambda_i\}$, $i = 1\cdots 4$, and the maximum height $H_{max}$. We acquired from a 3D satellite map the terrain and obstacles data inside the box. We denote $K$ as the total number of obstacles identified. These obstacles are characterized by a cylinder whose center coordinates on the ground are $C_k (\varphi_k, \lambda_k)$, $k = 1\cdots K$ with its avoidance radius $r_p$.

The formation shape and monitoring mission are used to find the UAV positions. Each UAV's position ($P_n$) must be considered to ensure safety among the group, but it can perform the mission well during real-time operation. The safe radius, kinematic constraints, of each UAV is also measured in this step. At the end of the preparation step, all information collected during the preparation phase is structured, encoded, and saved in an initiation file (Init_file).

### 4.2 Initialization

The initialization step is implemented after receiving all input data for the path planning problem. The tasks include initializing the working space, number of waypoints, parameters of the GEPSO algorithm, and assembling a random flight path to connect $P_s$ and $P_f$. In this case, the problem initiation means loading the Init_file into the global memory.

The following initial values of the flight mission problem are based on the cost functions built in section 2. The penalty coefficients of the individual cost functions have been carefully determined before integrating into the multi-objective problem (6). The initial value for the overall cost function ($\prod r(q)$) is infinite ($+\infty$). Paramaters related to UAVs, such as the location and constraint values of each UAV, have been predetermined according to the technical explanation and integrated into the computer. The initial UAVs' position, direction, and status coordinates are relative to the formation's reference point (centroid).

The GEPSO values' initialization can be manually entered into the computer depending on the input values. Choosing appropriate parameters, such as the number of individuals in the flock and the maximum number of runs (*N*), is an essential parameter that needs to be determined in advance. Choosing the appropriate parameters of GEPSO, such as $\omega, \alpha, \varphi, r$ and $\psi$ also needs to be considered.

### 4.3 Path generation

```
/* Path generation for the formation centroid: */
01 foreach i < (number_of_iterations) do
02 | foreach particle do
03 | | Initialize particles;           /* Eqs. (20,21) */
04 | foreach particle i do
05 | | Calculate cost value ∏^i r(q);   /* Eq. (6) */
06 | | If current cost < previous costs then
07 | | | Update new value for p_i^t ;
08 | | End If
09 | | Select a random value for p_i^t ;
10 | | Compute the value of global best p_g^t ;
11 | | Select a random value for v_rand ;
12 | | Select a random particle's best value p_rand ;
13 | | Update new result for ω_1 ;       /* Eq. (19) */
14 | | Calculate particle's velocity;   /* Eq. (17) */
15 | | Apply the velocity constriction;
16 | | Update new position;             /* Eq. (15) */
17 | | Apply the position constriction;
18 | | Check the safety of the path;    /* Eq. (10) */
19 | | Evaluate flight paths based on global cost function
         and safety cost function;
20 | | Update new results for p_i^t ;
21 | End
22 | Update p_g^t and ∏_safe ;
23 End
24 Save p_g^t and ∏_safe to the global memory;
25 Achieved the optimal path at the last iteration;
26 Generate trajectory for each UAV.
```

**H. 2**  *GEPSO path planning algorithm in psedo-code*

At this point, solving the path generation using the GEPSO algorithm for the formation can be described as a pseudo-programming code, as shown in H. 2.

## 5. Results

### 5.1 Mission set up

The assigned monitoring task in the implementation is to establish an optimal, safe flight trajectory using a UAV through a complex operational space. We used the Mission Planner (MP) app, a 3DR Solo drone, and a laptop ground control station to carry out the mission. MP is a UAV ground control software developed to prepare designed trajectories manually, set up command and flight missions. At the same time, MP monitors, intervenes, and directly controls the UAV while flying. Specifically, we use Google Satellite Maps (GST) on MP to gather initial information about the monitoring scene and surrounding working spaces. Finally, we can quickly determine the location and parameters of obstacles and then mark them on the map (H. 3).

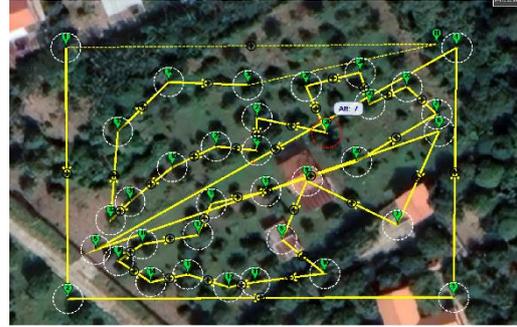

**H. 3**  *Working space acquisition*

The UAVs' mission is to monitor the fruit trees along their path. Their working space is inside a rectangular prism with its two opposite corners restricted by the GPS coordinates {12.233106, 109.114506}, {12.233563, 109.115220} and 30 meters high. The UAVs will take off at {12.233194, 109.114557} and land at {12.233411, 109.115187}. Dense obstacles with different radii and altitudes make the region to be complicated.

In the monitoring configuration, the positions of the UAVs in relation to the formation centroid are T1 = [0; 2; 0] m, T2 = [-2;-1; 0] m, and T3 = [2;-1; 0] m. The bounds of altitudes are set at $z_{max}$ = 5m and $z_{min}$ = 4 m, respectively. The UAVs fly within the relative distance range to the trees limited by [0,5, 4] m.

The number of individuals and iterations in the proposed algorithm are both 100. The GEPSO parameters are initialized, as illustrated in Table 1.

**Table 1.** GEPSO parameters selected

| Symbol | Value | Symbol | Value | Symbol | Value |
|---|---|---|---|---|---|
| $\omega_1, \omega_2$ | 0.5 | $\alpha_2, \alpha_3$ | 2.0 | $r_1^i, r_2^i$ | 2.0 |
| $\omega_3, \omega_4$ | 0.8 | $\varphi_1, \varphi_3$ | 2.0 | $r_3^i$ | 1.5 |
| $\omega_5$ | 0.9 | $\varphi_2$ | 3.0 | $r_4^i$ | 1.5 |
| $\alpha_1$ | 4.5 | $\varphi_4$ | 2.0 | $\psi$ | 0.9 |

### 5.2 Experiments and comparisons

The implementation aims to find an optimal and safe path for UAV formation using the proposed algorithm.

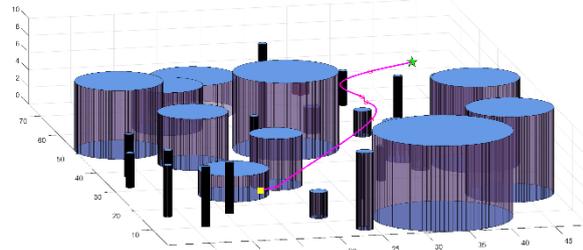

**H. 4**    *Formation centroid 3D-path generated*

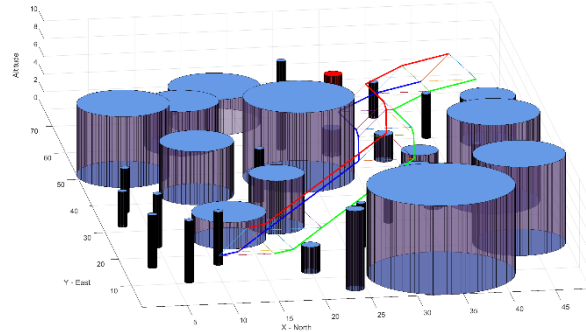

**H. 5**    *Generated path for the three UAVs in 3D space*

The 3D trajectories generated for the formation centroid and the three UAVs in H. 4 and H. 5 show that the generated flight path algorithms can reach the target in harsh working conditions. In H.5, we observe that the formation shape is preserved through triangular connections at each waypoint. The projections of planning paths to show the flyability, safety and feasibility paths are illustrated in H. 6 and H. 7. The two most important criteria are (1) the shortest flight path and (2) safety when not intersecting treacherous areas of obstacles.

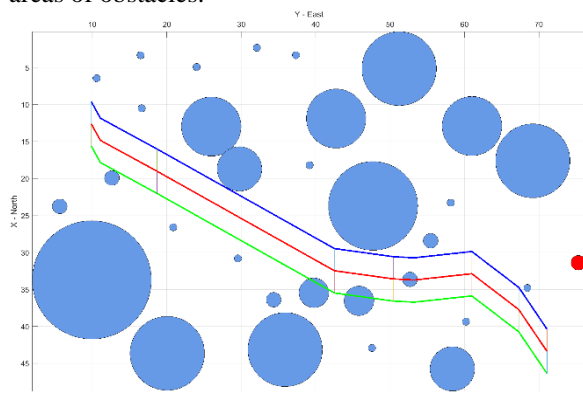

**H. 6**    *Generated path in the ground (xy plane)*

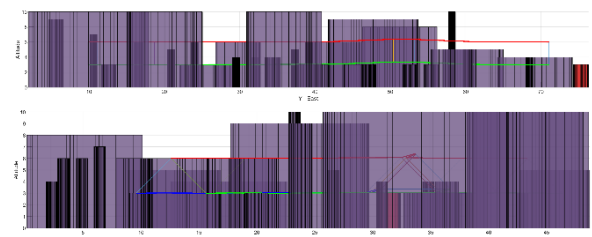

**H. 7**    *Generated paths in vertical planes*

Furthermore, we apply the generated paths for the three UAVs on Mission Planner employing its corresponding satellite map, as clarified in H. 8. The results show that the planned paths are all flyable for the UAVs to operate with slight and smooth curvature. The results also show that all the generated paths for the UAVs are capable of avoiding densely arranged obstacles in the complicated environment, ensuring safety for the UAV throughout the journey. Accordingly, the results have evaluated the feasibility of the proposed algorithm.

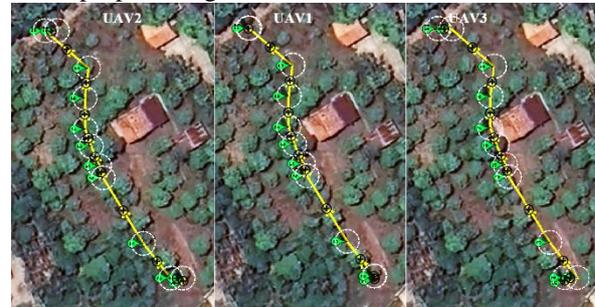

**H. 8**    *Trajectory commands for UAVs' controllers*

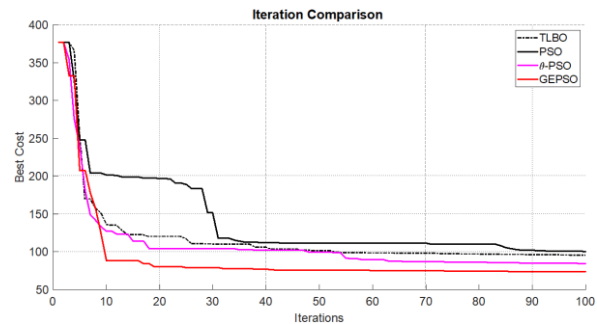

**H. 9**    *Convergence comparison*

For validation, we compared the performance of GEPSO with θ-PSO [10] and TLBO [8] planning algorithms, as illustrated in H. 9, which is the average of 30 consecutive runs. H. 9 shows the changing values of the quality cost function over iterations. It can be seen that the GEPSO algorithm converges to the smallest value (73.35) and is more stable in a lesser number of iterations (65) in comparison with PSO (93.03, 95), θ-PSO (83.78, 82) and TLBO (96.67, 121).

### 6. Conclusion

This paper has introduced a new approach for a multi-copter UAV's route planning algorithm in surface monitoring applications. The generalized particle swarm optimization algorithm is utilized to provide command paths for the multi-copter UAVs constructed in a formation. To improve operational safety and mission efficiency, the GEPSO then incorporated constraints to make it a multi-objective optimization algorithm for the path planning process. The study also suggested satellite maps to enhance their practical application. Implementation and comparisons have been made to demonstrate the path planning algorithm's performance in complicated working environments. Future works will focus on implementing the proposed approach into actual drones in real and practical situations.

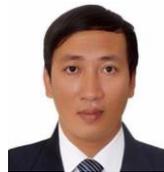

**Van Truong Hoang**, born in 1978, received the M.Eng. degree in electronic engineering from La Trobe University, Melbourne, Australia, in 2010 and the Ph.D. degree in control and automation from the University of Technology, Sydney, Australia in 2019. He is now a lecturer of Faculty of Missile and Ship-gun, Naval Academy, Vietnam. His research interests include control and path planning for multiple unmanned aerial vehicles coordinated in a formation and its application in flying machines.